\documentclass[review]{elsarticle}
\graphicspath{ {./figures/} }
\usepackage{hyperref}
\usepackage{float}
\usepackage{verbatim} 
\usepackage{apalike}
\restylefloat{figure}
\floatstyle{plaintop} 
\restylefloat{table}

\usepackage{amsmath} 
\usepackage{amssymb}
\usepackage{array} 
\usepackage{adjustbox}
\usepackage{algorithm}
\usepackage{algpseudocode}
\usepackage{geometry}
\usepackage{subcaption}
\usepackage{graphicx}
\usepackage{float}
\usepackage{multirow}

\usepackage{algorithm}%
\usepackage{algorithmicx}%
\usepackage{float}
\usepackage{hyperref}
\usepackage{verbatim} 
\floatstyle{plaintop} 
\restylefloat{table}
\usepackage{array} 
\usepackage{adjustbox}
\usepackage{geometry}
\usepackage{subcaption}
\usepackage{graphicx}
\usepackage{orcidlink} 
\usepackage{cellspace} 
\usepackage{booktabs}

\newcommand{\maxtablewidthh}{0.7\textwidth} 
\newcommand{\maxtablewidth}{0.6\textwidth} 

\biboptions{authoryear}

\begin{document}

	\begin{frontmatter}
		

		\title{Knowledge Distillation on Spatial-Temporal Graph Convolutional Network for Traffic Prediction}
		
		\author[label1]{Mohammad Izadi}
		\ead{m.izadi@ec.iut.ac.ir}
		
		\author[label1]{Mehran Safayani \corref{cor1}}
		\ead{safayani@iut.ac.ir}
		
		\author[label1]{Abdolreza Mirzaei }
		\ead{mirzaei@iut.ac.ir}
		
		\cortext[cor1]{Corresponding author.}
		\address[label1]{Department of Electrical and Computer Engineering, Isfahan University of Technology, Isfahan 84156-83111, Iran}

		\begin{abstract}
			Efficient real-time traffic prediction is crucial for reducing transportation time. To predict traffic conditions, we employ a spatio-temporal graph neural network (ST-GNN) to model our real-time traffic data as temporal graphs. Despite its capabilities, it often encounters challenges in delivering efficient real-time predictions for real-world traffic data. Recognizing the significance of timely prediction due to the dynamic nature of real-time data, we employ knowledge distillation (KD) as a solution to enhance the execution time of ST-GNNs for traffic prediction. In this paper, We introduce a cost function designed to train a network with fewer parameters (the student) using distilled data from a complex network (the teacher) while maintaining its accuracy close to that of the teacher. We use knowledge distillation, incorporating spatial-temporal correlations from the teacher network to enable the student to learn the complex patterns perceived by the teacher. However, a challenge arises in determining the student network architecture rather than considering it inadvertently. To address this challenge, we propose an algorithm that utilizes the cost function to calculate pruning scores, addressing small network architecture search issues, and jointly fine-tunes the network resulting from each pruning stage using KD. Ultimately, we evaluate our proposed ideas on two real-world datasets, PeMSD7 and PeMSD8. The results indicate that our method can maintain the student's accuracy close to that of the teacher, even with the retention of only $3\%$ of network parameters.
		\end{abstract}
		
		
	\end{frontmatter}
\section{Introduction}
Traffic prediction is a critical task in urban planning, logistics, and transportation management \cite{ref1,ref21}. The data necessary for accurate prediction is typically sourced from monitoring stations strategically positioned throughout urban areas. These stations continuously collect a plethora of traffic-related parameters, including vehicle speed, traffic density, and flow rates, at regular intervals. This data is then structured into temporal graphs, with each node representing the traffic conditions observed at a specific monitoring station, and the edges indicating the connections between these stations \cite{ref1, ref43}. To leverage this wealth of data for predictive purposes, spatio-temporal graph neural networks (ST-GNNs) have emerged as a promising approach. These networks have the capability to analyze the temporal evolution of traffic patterns and make accurate predictions regarding future traffic conditions for each monitoring station within desired time intervals. However, a significant challenge arises when dealing with a large number of nodes, as it can lead to computationally expensive operations and demand substantial hardware resources \cite{ref1}. \\	
In response to this challenge, our research focuses on enhancing the execution time of ST-GNNs through the application of an integrated algorithm that combines knowledge distillation and network pruning \cite{ref2,ref24,ref25,ref27,ref28, ref40, ref41, ref42, ref43, ref8}. \\
Previous works have not implemented an integrated algorithm to address the structure search for student networks by considering the capability of learning by filters using knowledge distillation. These studies typically consider the student structure static, or those that do address structure search for finding student structures do not mention using knowledge distillation to address the capability of filters to learn from the teacher. Instead, they simply use task-specific cost functions. In contrast, we address both aspects within a single algorithm, integrating them seamlessly in one training phase. Our approach uses knowledge distillation to train a simplified "student" network with fewer parameters, helping it to understand complex spatial and temporal data by learning from a more complex "teacher" network. To refine the student network's architecture, we introduce a pruning algorithm that scores the capability of student filters in learning this information. Insignificant parameters are eliminated, and after each pruning stage, we fine-tune the network using knowledge distillation. This iterative process enhances the student network's efficiency and learning capability, leading to improvements in execution time without compromising accuracy \cite{ref2,ref17,ref18,ref19,ref16}. By integrating pruning and knowledge distillation, our method effectively addresses the dual challenge of structuring student networks and enhancing their ability to learn complex data from the teacher model, resulting in a streamlined and effective training process.\\
The evaluation of our proposed approach is conducted using real-world traffic datasets, PeMSD7 and PeMSD8. Our results demonstrate that the student network trained using our cost function outperforms previous approaches in terms of both knowledge distillation effectiveness and execution time reduction. By jointly applying knowledge distillation and pruning, we assess the competency of network parameters, ensuring efficient model compression without sacrificing predictive accuracy.\\
The structure of the paper is organized as follows: Section 2 explores related works on knowledge distillation, network pruning, and spatio-temporal graph neural networks. Section 3 introduces a ST-GNN model and the proposed knowledge distillation methods. It also discusses a pruning algorithm designed to address challenges in the student network architecture. Section 4 evaluates the performance of various cost functions on the PeMSD7 and PeMSD8 datasets, presenting ablation studies and analyzing their results. Finally, Section 5 outlines future works and draws conclusions.	
\\

\section{Related Works}
\label{title_page}

\subsection{Traffic Prediction and Spatio-Temporal Graph Neural Networks}

Fast and accurate urban traffic prediction is vital for traffic control and management, especially for medium and long-term forecasts. Traditional methods often struggle with complexities in traffic flow and neglect spatial and temporal dependencies. To address this, deep networks capable of handling such data have been proposed.\\
In \cite{ref6}, the significance of traffic network prediction and its applications, such as network monitoring, is discussed. Various recurrent neural network architectures, including standard RNN, LSTM, and GRU, are explored for traffic prediction. However, these models suffer from high execution times and difficulties in capturing long-range dependencies due to numerous parameters.\\
Bidirectional LSTM models are found to outperform unidirectional LSTM models in \cite{ref7}, offering improved accuracy. Despite this, the high computational cost remains a significant drawback, making these models less suitable for real-time applications.\\
Spatio-temporal graph neural networks (ST-GNNs) are introduced to address these limitations by utilizing graph convolution layers for spatial processing and convolution layers for capturing temporal correlations. This approach offers reduced execution time and comparable or better accuracy.\\ In \cite{ref5}, the attentional spatio-temporal graph convolutional network (ASTGCN) model effectively predicts traffic flow by leveraging spatial-temporal attention mechanisms and convolution layers. However, the model complexity and training time can still be high.\\
The spatio-temporal graph convolutional networks (STGCN) framework presented in \cite{ref1} shows enhanced training efficiency and captures comprehensive spatial-temporal correlations, outperforming other methods on real traffic datasets. Yet, it still faces challenges in scaling to very large networks and handling diverse urban environments.\\
Recent surveys such as \cite{ref43} and \cite{ref44} provide comprehensive overviews of advancements in this field, highlighting the progress and persistent challenges such as handling heterogeneous data sources and improving model interpretability.
\subsection{Knowledge Distillation}
\textit{Knowledge Distillation}, or \textit{Teacher-Student Learning}, introduced in \cite{ref2} in 2015, transfers knowledge from a complex "teacher" network to a smaller "student" network. This alignment of behavior and predictions improves student performance and reduces computational resources.\\
Further research extends this to graph and spatio-temporal networks. In \cite{ref9}, a novel approach for knowledge transfer from a GCN teacher model to a GCN student model is introduced, leading to a smaller model with improved execution time, particularly suited for dynamic graph models. However, this method may struggle with the heterogeneity of traffic data.\\
In \cite{ref10}, the objective is to train a simpler network through knowledge distillation from multiple GCN teacher models with different tasks, extending to distill spatial graph information from hidden convolutional layers. This approach empowers the student model to excel in various tasks without additional labeled data, but the integration of heterogeneous knowledge remains challenging.\\
In \cite{ref11}, knowledge distillation is extended to ST-GNNs for modeling spatial and temporal data in human body position videos, utilizing various techniques, including minimizing loss, leveraging spatial-temporal relations, and using the gradient rejuvenation method to optimize the student model. Despite its success, adapting this approach to the diverse and dynamic nature of urban traffic data poses significant challenges.
\subsection{Pruning and Fine-Tuning}
In 1990, neural network pruning was first introduced by \cite{ref3}, offering a method to identify connections safe for pruning using the Hessian matrix. This reduces parameters in large networks, mitigating overfitting and resource requirements, crucial for deployment on constrained devices. Pruning has since become essential for compressing networks across diverse applications.\\ 
In 2019, \cite{ref8} proposed a method to estimate layer and weight importance, facilitating the removal of unimportant weights and network size reduction. Iterative pruning and retraining stages improve performance and reduce model complexity. However, maintaining performance while significantly reducing model size remains a persistent challenge.

\section{Proposed Methods}
Traffic prediction, involves predicting future values of key traffic conditions (e.g., speed or traffic flow) for the next \(h\) time steps based on \(M\) prior traffic observations.
In this paper, we model the traffic network as graphs defined across time steps. At time step \(t\), data from \(N\) monitoring stations is represented as a graph \(G_t = (V_t, E, W)\), where \(V_t \in \mathbb{R}^N\) contains features from \(N\) nodes (e.g., road segments) with speed as the chosen criterion. The edge set \(E\) depicts connections between stations, defined based on distance criteria, reflecting the influence of stations on each other. Weights \(W \in \mathbb{R}^{N \times N}\) in the weighted adjacency matrix signify the relationships between stations, often determined by spatial disparities.\\
The ST-GCN\cite{ref1}, designed for traffic prediction, processes spatial and temporal data. Graph convolution layers capture spatial relationships and node features over time. Convolutional layers model temporal correlations, focusing on sequential patterns and changes between graphs. The architecture comprises two blocks with temporal and spatial layers, and an output block transforms information into the final output.  The teacher and student networks share an identical number of blocks and layers. Any variations in parameters result from differences in channel numbers within the temporal and spatial layers of the hidden layer.
The complex ST-GCN teacher network achieves high accuracy with 333,604 and 296,426 parameters for PeMSD7 and PeMSD8 datasets ( Table \ref{tab:3-1} and Figure \ref{fig:loss} ). Despite accuracy, its high parameter count results in significant computational cost ( see Table \ref{tab:3-1}, first row ).
A proposed lighter model, the student network ( see Table \ref{tab:3-1}, third row ) is a scaled-down version of the teacher network. With 10,144 and 7,766 parameters for PeMSD7 and PeMSD8, it maintains accuracy while reducing complexity and computational requirements. This lightweight solution is efficient for traffic prediction, especially in resource-constrained scenarios.
\subsection{Knowledge Distillation}
\begin{figure*}[!h]
	\centering
	
	\begin{adjustbox}{center, margin=0, width=\maxtablewidth}
		\includegraphics[width=1\textwidth]{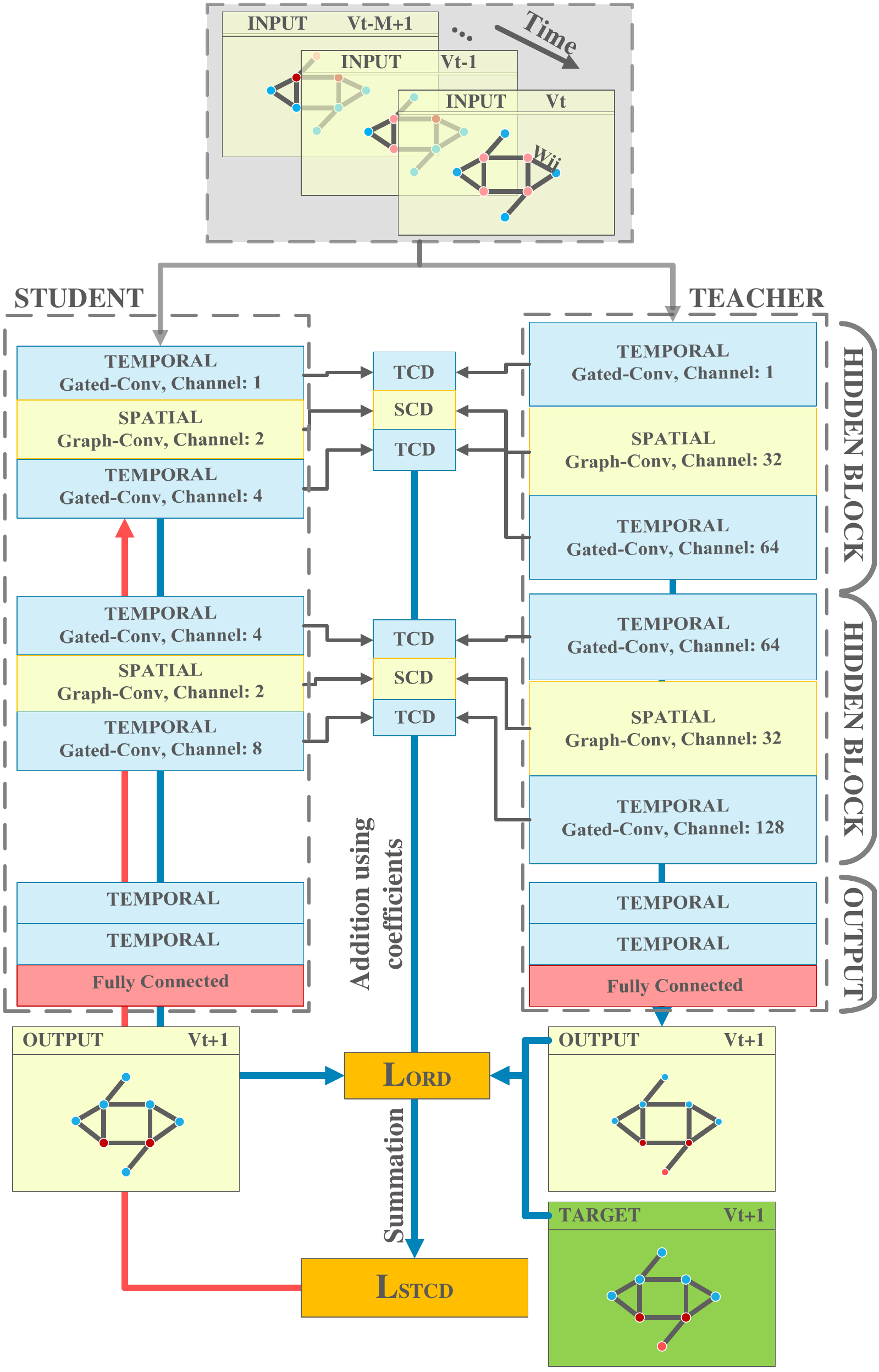}
	\end{adjustbox}
	\caption{The illustration features both our student and teacher models, demonstrating the application of our cost functions to the spatio-temporal graph convolutional network (ST-GxCN)}
	\label{fig:loss}
\end{figure*}

In this paper, we use offline distillation, wherein the student distills information from the teacher network after the teacher network has been completely trained\cite{ref30,ref31,ref32}. We utilized two knowledge distillation techniques. The initial approach, known as response-based distillation, involves transmitting crucial information from the teacher network's outputs to the student network. This leads to the development of a simpler and faster model while simultaneously preserving performance. The second method, feature-based knowledge distillation, focuses on hidden layer`s knowledge, encompassing spatial and temporal correlations among graph nodes (Figure \ref{fig:loss}).

\begin{table*}[!h]%
	\centering
	\caption{Network information for knowledge distillation and pruning}
	\label{tab:3-1}	
	\begin{adjustbox}{center, margin=0, width=\maxtablewidth}
		\begin{tabular}{|Sc|Sc|Sc|Sc|Sc|}
			\hline
			\multicolumn{5}{|Sc|}{PeMSD7} \\
			\hline
			Models & Parameters & Hidden Blocks Channel & Test Time (s) & FLOPS \\
			\hline
			Teacher  & 333,604 & [1, 32, 64][64, 32, 128] & 3.423 & 49,889,172,087\\
			\hline
			Pruning Base Model  & 48,628 & [1, 8, 16][16, 8, 32] & 1.069 & 9,113,934,711\\
			\hline
			Student  & 10,144 & [1, 2, 4][4, 2, 8] & 0.547 & 1,726,990,455\\
			\hline
		\end{tabular}
			\end{adjustbox}
		\quad
			\begin{adjustbox}{center, margin=0, width=\maxtablewidth}
		\begin{tabular}{|Sc|Sc|Sc|Sc|Sc|}
			\hline
			\multicolumn{4}{Sc|}{PeMSD8} \\
			\hline
			Models & Parameters & Hidden Blocks Channel & Test Time (s) & FLOPS \\
			\hline
			Teacher  &296,426 & [1, 32, 64][64, 32, 128] & 2.556 & 40,636,466,453\\
			\hline
			Pruning Base Model  &39,290 & [1, 8, 16][16, 8, 32] & 0.810 & 5,659,617,749\\
			\hline
			Student  & 7,766 & [1, 2, 4][4, 2, 8] & 0.441 & 1,003,700,933\\
			\hline
		\end{tabular}
	\end{adjustbox}
\end{table*}

\subsubsection{Response-based Distillation}

Leveraging the L2 and Kullback-Leibler (KL) divergence metrics, these functions measure the differences between the teacher and student network outputs. The student network endeavors to enhance its accuracy compared to the teacher by minimizing these error functions. Equations \eqref{eq:3-1} and \eqref{eq:3-2} employ the L2 and KL error functions to quantify the difference between the outputs of the teacher and student networks for each node.\\
\begin{equation}
	\label{eq:kl}
	\forall b \in B, \forall i \in N: \quad \text{KL}(y^{s}_{bi}, y^{t}_{bi}) = y^{t}_{bi} \log\left(\frac{y^{t}_{bi}}{y^{s}_{bi}}\right)
\end{equation}
\begin{equation}
	\label{eq:3-1}
	\forall b \in B, \forall i \in N: \quad L_\mathrm{RD(KL)bi} = \beta \cdot \text{KL}(y^{s}_{bi}, y^{t}_{bi}) + (1 - \beta) \cdot \| y^{s}_{bi} - T_{bi} \|_2
\end{equation}
\begin{equation}
	\label{eq:3-2}
	\forall b \in B, \forall i \in N: \quad L_\mathrm{RD(L2)bi} = \beta \cdot \| y^{s}_{bi} - y^{t}_{bi} \|_2 + (1 - \beta) \cdot \| y^{s}_{bi} - T_{bi} \|_2
\end{equation}

\mbox{}\\
In these equations, $|| \cdot ||_2$ denotes the L2 norm. The variables $N$, $y^{s}$, and $y^{t}$ represent the number of nodes, outputs of the student network, and outputs of the teacher network, respectively. KL is the Kullback-Leibler divergence metric (Equation~\eqref{eq:kl}), $T_{bi}$ is the target data value for batch $b$, and $\beta$ is an adjustable coefficient. These cost functions consider both the teacher and target data jointly. \mbox{}\\
However, we propose an alternative approach described by Equations \eqref{eq:3-3} to \eqref{eq:3-5} to determine when to use teacher predictions versus target data during student training. This decision hinges on the difference between teacher predictions and target data. If this difference exceeds a threshold ($\alpha_1$), it indicates that even the more complex teacher model struggles to make an accurate prediction. In such cases, the simpler student model, with fewer parameters, is likely to face similar difficulties. Therefore, it is beneficial for the student to follow the teacher's prediction, which tends to be smoother and less affected by noise. Otherwise, we use the target data directly. This method helps mitigate noise in the target data while leveraging the teacher's more refined predictions to guide the student's learning process effectively.
\begin{equation}
	\label{eq:3-3}
	\forall b \in B, \forall i \in \mathbb{N} : D^{t}_{bi} = \left| {y^{t}_{bi}} - T_{bi} \right|
\end{equation}
\begin{equation}
	\label{eq:3-4}
	D^{t}_{bi} = \frac{D^{t}_{bi} - \min(D^{t}_{b})}{\max(D^{t}_{b}) - \min(D^{t}_{b})}
\end{equation}
\begin{equation}
	\label{eq:3-5}
	L_\mathrm{ORD} += \sum_{b \in B} \sum_{i \in N} \left\{
	\begin{array}{ll}
		\|y^{s}_{bi} - T_{bi}\|_2 & \text{ : } D^{t}_{bi} \leq \alpha_1 \\
		\|y^{s}_{bi} - y^{t}_{bi}\|_2 & \text{ : } D^{t}_{bi} > \alpha_1
	\end{array}
	\right.
\end{equation}
\mbox{}\\
In these equations, $D^{t}_{bi}$ represents the absolute differences between teacher predictions and target data for each node in batch $b$. After normalization (Equation \eqref{eq:3-4}), each element is compared with the threshold $\alpha_1$ to identify potentially noisy data. The loss function $L_{ORD}$ is then computed by summing these values for each node across all batches. You can observe the representation of this cost function in Figure \ref{fig:loss}.
\subsubsection{Feature-based Knowledge Distillation in the Hidden Layers}
Knowledge distillation from hidden layers simplifies the training process for deep and complex neural networks, making it less intricate and time-consuming. This method enables the training of simpler models that preserve meaningful information from the data and leverage the hidden knowledge of more complex models. In the following subsections, we elaborate on our cost functions designed to capture both spatial and temporal correlations between the teacher and student networks. These cost functions aim to align the output of corresponding layers, fostering a close relationship in both spatial and temporal aspects.
\mbox{} \\
\textbf{Temporal Correlation Distillation} We introduce the cost function \(L_{\text{TCD}}\) (as defined in equations \eqref{eq:3-6} and \eqref{eq:3-7} and illustrated in  Figure \ref{fig:loss} ) to ensure that the output of temporal layers in the student network closely aligns with the corresponding layers in the teacher network. The objective of \(L_{\text{TCD}}\) is to enable the temporal layers in the student network, which have fewer parameters compared to equivalent layers in the teacher network, to perform similarly.
\begin{align}
	\label{eq:3-6}
	TCD_{bnij} &= \frac{1}{C} \sum_{c=1}^{C} \left| F_{binc} - F_{bjnc} \right|, \quad TCD \in \mathbb{R}^{B \times N \times T \times T}
\end{align}

\begin{equation}
	\label{eq:3-7}
	L_{\text{TCD}} = \frac{1}{{B \cdot N \cdot \binom{T}{2}}} \sum_{b=1}^{B} \sum_{n=1}^{N} \sum_{{\substack{i, j \\ j>i}}}^{T} \|TCD_{bnij}^s - TCD_{bnij}^t\|_2
\end{equation}
In these equations, \(F \in \mathbb{R}^{B \times T \times N \times C}\), where \(B\) is the batch size, \(T\) is the number of time steps, \(N\) is the number of graph nodes, and \(C\) is the number of channels. Each element of the \(TC\) tensor represents the absolute difference in features of a node at two time steps \(i\) and \(j\). The cost function \(L_{\text{TCD}}\) addresses the difference in values for each node at two time steps. These errors are calculated using Equation~\eqref{eq:3-6} for both the teacher network and the student network. In the equation \(\binom{T}{2}\), it accounts for the selection of any two time steps \(i\) and \(j\) out of \(T\) due to normalization. Finally, the difference between these two sets of Equations \eqref{eq:3-7} yields the \(L_{\text{TCD}}\) cost function.
\mbox{} \\
\textbf{Spatial Correlation Distillation} Assuming a resemblance between the output of a spatial layer and the temporal layers, the cost function computes pairwise differences in values for all nodes at each time step in both the student and teacher networks (see Equation~\eqref{eq:3-8} ). These values are calculated for all spatial layers in both networks, and their pairwise differences are determined. To ensure a close alignment of the output from spatial layers in the student network with the corresponding layers in the teacher network, we introduce the cost function \(L_{\text{SCD}}\), as defined in Equation~\eqref{eq:3-9} and illustrated in Figure \ref{fig:loss}.\\
\begin{equation}
	\label{eq:3-8}
	\begin{split}
		SCD_{btij} &= \frac{1}{C} \sum_{c=1}^{C} \left| F_{btic} - F_{btjc} \right|, \quad SCD \in \mathbb{R}^{B \times T \times N \times N}
	\end{split}
\end{equation}
\begin{equation}
	\label{eq:3-9}
	L_\mathrm{SCD} = \frac{1}{{B \cdot T \cdot \binom{N}{2}}} \sum_{b=1}^{B} \sum_{t=1}^{T} \sum_{{\substack{i, j \\ j>i}}}^{N} \|SCD_{btij}^s - SCD_{btij}^t\|_2
\end{equation}
Each element of the \(SC\) tensor represents the absolute difference in features of a node with another node at time step \(t\). This tensor is computed for both the teacher and student networks (Equation~\eqref{eq:3-8}). The \(L_{\text{SCD}}\) cost function, calculated as the difference between these two vectors, is obtained from Equation~\eqref{eq:3-9}.
\subsubsection{Space-Time Cost Function}
Through the integration of three distillation components, we construct a comprehensive cost function aimed at distilling relevant information pertaining to relationships within both temporal and spatial layers. The derived cost function is defined in Equation~\eqref{eq:3-10} and depicted in Figure \ref{fig:loss}.
\begin{align}
	\label{eq:3-10}
	L_{\text{STCD}} &= L_{\text{ORD}} + \alpha_3 \cdot (\alpha_2 \cdot L_{\text{SCD}} + (1 - \alpha_2) \cdot L_{\text{TCD}})
\end{align}
\mbox{} \\
In this equation, the cost function \(L_{\text{STCD}}\) is a composite of three components. First we use \(L_{\text{ORD}}\) to address response distillation. In the feature-based area we use \(L_\mathrm{SCD}\) and \(L_\mathrm{SCD}\) to deal with spatial and temporal blocks in hidden layers. The parameters \(\alpha_2\) and \(\alpha_3\) allow adjusting the importance between temporal and spatial distillations and the significance given to distilling hidden layers in the cost function.
\subsection{Pruning}
In this subsection, we present an algorithm that employs the \(L_{\text{STCD}}\) cost function to train the student network. Additionally, it extracts the architecture of the student network by pruning insignificant parameters from the teacher network. The significance of each neuron is evaluated based on its influence on the network's accuracy and its capacity to learn features extracted from the teacher. As outlined in the method presented by \cite{ref8}, Equation~\eqref{eq:3-14} is utilized to establish a score \(I\) for a set of parameters labeled as $w_s$, similar to a convolutional filter. It is defined as a contribution to group sparsity in Equation~\eqref{eq:3-14}:
\begin{equation}
	\label{eq:3-14}
	\text{I}_S^{(1)}(W, B) \overset{\triangle}{=}  \sum_{s \in S} \text{I}_s^{(1)}(W, B) = \sum_{s \in S} \left(\sum_{i=1}^{B} (g_{s,i} \, w_{s,i})^2\right)
\end{equation}
In this paper, Equation~\eqref{eq:3-14} is used to obtain the importance score of parameters. However, instead of using the gradients and weights obtained from training the network in the standard form, the network is trained using cost function defined in the section 3.2. The importance score obtained from the gradients and weights of the network, this time, not only indicates the importance of the parameter in the output but also reflects the learning ability of the knowledge perceived and extracted by the teacher. The proposed method is outlined in Algorithm 1. The algorithm defines a mask matrix \( M \) to selectively retain or discard parameters in the network, with values of 1 or 0, respectively. An importance matrix \texttt{KDIS} is initialized to calculate the importance score of each parameter. The algorithm specifies minibatch intervals for pruning steps and the percentage of parameters to be pruned at each step. During each minibatch, the model undergoes fine-tuning using the \(L_{\text{STCD}}\) loss function (line 11), where gradients and importance scores (\texttt{KDIS}) are computed based on the network weights. Knowledge distillation (\textit{KD}) is applied to calculate gradients and weights, denoted as \texttt{KDIS} instead of \textit{I}. After a specified minibatch count in line 5, importance values are averaged, and parameters are pruned in each layer according to the specified percentage (lines 13 to 19). The pruned network is then fine-tuned over multiple epochs (lines 24 and 25). The resulting mask matrix \( M \) contains values of zero (removed parameters) and one (remaining parameters). Multiplying this matrix by the weight matrix of the student network's base architecture yields the final pruned network.
\begin{algorithm}[!h]
	
	\setlength{\textwidth}{\linewidth}
	\setlength{\lineskip}{0.6ex}
	\setlength{\lineskiplimit}{\maxdimen}
	
	\caption{Jointly KD-Pruning Algorithm}
	\begin{algorithmic}[1]
		\State \textbf{Input:} \texttt{base\_model} - Pre-trained model
		\State \textbf{Output:} Pruned mask \texttt{M}
		\State Initialize:
		\State \texttt{minibatch\_counter} $\gets$ 0
		\State \texttt{pruning\_minibatch} $\gets$ \textit{your value here}
		\State \texttt{KDIS} $\gets$ Zero matrix with the shape of hidden block weights
		\State \texttt{M} $\gets$ 1 matrix with the shape of hidden block weights
		\State \texttt{pruning\_percentage} $\gets n\%$

		\For{\texttt{each minibatch}}
		\State \texttt{minibatch\_counter} $\gets$ \texttt{minibatch\_counter} $+$ 1
		\State \texttt{grads, weights} $\gets$ \texttt{Finetune(base\_model, \(L_{\text{STCD}}\), M)}
		\State \texttt{KDIS} $\gets$ \texttt{KDIS} $+$ \texttt{compute\_KDIS(grads, weights)}
		\If{\texttt{minibatch\_counter} $==$ \texttt{pruning\_minibatch}}
		\State \texttt{minibatch\_counter} $\gets$ 0
		\State \texttt{KDIS} $\gets$ \texttt{KDIS / pruning\_minibatch}
		\Comment{Update pruning decision}
		\For{\texttt{each layer \textbf{in} layers}}
		\State \texttt{sort(KDIS[l], ASC)}
		\For{\texttt{each index \textbf{in} first\_Npercent\_indexes(KDIS)}}
		\State \texttt{M[l][p]} $\gets$ 0 \Comment{Prune N\% of parameters with the lowest KDIS}
		\EndFor
		\EndFor
		\EndIf
		\EndFor
		\For{\texttt{several epochs}}
		\State \texttt{Finetune(base\_model, ST-GCN, M)}
		\EndFor
		
		\State \textbf{return} \texttt{M}
	\end{algorithmic}
\end{algorithm}

	\section{Experiments and results}
\begin{figure*}[!h]
	\centering
	\begin{adjustbox}{center, margin=0, width=\maxtablewidth}
		\includegraphics[width=1.6\textwidth]{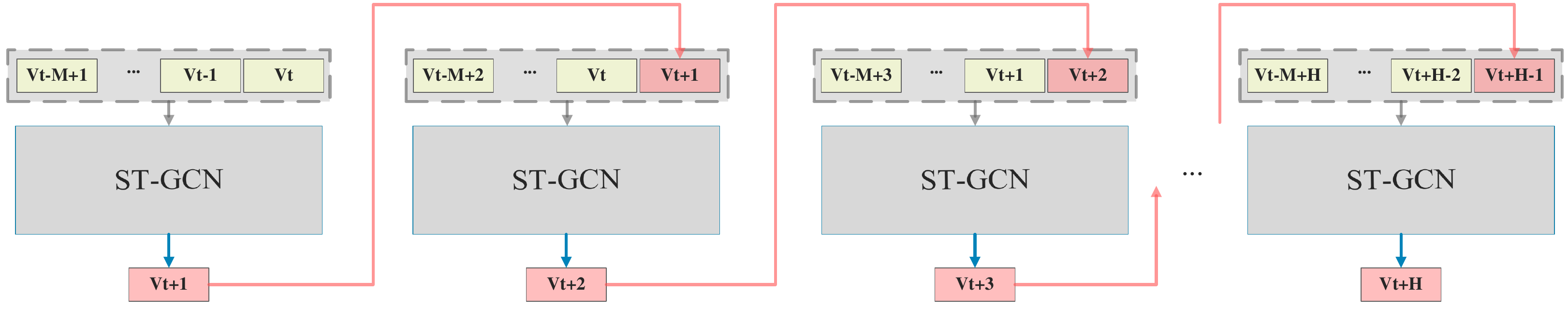}
	\end{adjustbox}
	\caption[Sequence of ST-GCN predictions for h future time steps]{Sequence of ST-GCN predictions for future time steps. In each step, the output of the current state is used as the last input graph to predict the next timestep.}
	\label{fig:prediction}
\end{figure*}

The results of traffic prediction in all tables are reported for the next 15, 30, and 45 minutes. The output values and calculated errors for these three time units are reported using three error functions: MAPE (Mean Absolute Percentage Error), MAE (Mean Absolute Error), and RMSE (Root Mean Square Error). The execution time reported in the tables are based on a batch of 1140 data and average over 100 runs. We thoroughly evaluate our model by testing it on two real traffic datasets: PeMSD7, which includes 228 nodes, and PeMSD8, with 170 nodes. 
\textbf{PeMSD7} is collected by over 39,000 sensor stations located throughout the main urban areas of the California state freeway system \cite{ref1}. In this paper, we randomly select an average scale from Region 7 of California, consisting of 228 stations, labeled as PeMSD7. \textbf{PeMSD8} is similar to PeMSD7, with the difference that it includes traffic data for the city of San Bernardino from July to August 2016, collected from 170 monitoring stations on 8 streets at 5-minute intervals. All presented results were generated using a TI 1080 GTX graphics card. We consistently used 12 historical timesteps \(M\) to predict 9 future timesteps \(H\). Our model predicts these 9 timesteps sequentially, as illustrated in Figure \ref{fig:prediction}. In this figure, each prediction (e.g., \(V_{t+1}\)) is utilized as the last graph in the input series to forecast the next timestep (\(V_{t+2}\)). This sequential approach enables us to predict \(h\) future timesteps. With data collected at 5-minute intervals, selecting 9 time units for prediction allows us to report results for the next 15, 30, and 45 minutes. Throughout all runs, the learning rate decreases by a factor of 0.7 after 5 epochs.

\subsection{Hyperparameters}
You can refer to Tables \ref{tab:4-1} for the learning rate, batch size, and hyperparameters related to knowledge distillation. In the context of the provided equations, the "Models" column enumerates various knowledge distillation approaches evaluated for PeMSD7 and PeMSD8 datasets. The hyperparameters $\alpha_1, \alpha_2, \alpha_3, \alpha,$ and $\beta$ are pivotal in the knowledge distillation process, as defined in equations \eqref{eq:3-5}, \eqref{eq:3-10}, \eqref{eq:3-1}, and \eqref{eq:3-2}. The hyperparameters for the proposed pruning algorithm are presented in Table \ref{tab:4-2}. The table categorizes hyperparameters based on the percentage of pruning applied, as well as batch size, learning rate, and three alpha values ($\alpha_1$, $\alpha_2$, and $\alpha_3$), specific to the PeMSD7 and PeMSD8 datasets. For instance, in the PeMSD7 row with a pruning percentage of $97\%$, only $3\%$ of the model parameters are retained. The associated hyperparameters include a batch size of 25, a learning rate of $1 \times 10^{-3}$, and alpha values $\alpha_1 = 0.099$, $\alpha_2 = 0.091$, and $\alpha_3 = 0.531$. 
\begin{table*}[!h]%
	\centering
	\caption{Hyperparameters of the proposed knowledge distillation loss functions}
	\label{tab:4-1}
	\begin{adjustbox}{center, margin=0, width=\maxtablewidth}
		\begin{tabular}{|Sc|Sc|Sc|Sc|Sc|Sc|Sc|Sc|}
			\hline
			\multicolumn{8}{|Sc|}{PeMSD7} \\
			\hline
			Models & Batch Size & Learning Rate & $\alpha$1 & $\alpha$2 & $\alpha$3 & $\alpha$ & $\beta$ \\
			\hline
			\(L_{\text{RD(L2)}}\) & 50 & 1E-03 & - & - & - & - & 0.045 \\
			\hline
			\(L_{\text{RD(KL)}}\) & 50 & 1E-03 & - & - & - & - & 0.007 \\
			\hline
			\(L_{\text{SKD}}\) & 16 & 1E-03 & - & - & - & 0.333 & - \\
			\hline
			\(L_{\text{ORD}}\)(ours) & 50 & 1E-03 & 0.593 & - & - & - & - \\
			\hline
			\(L_{\text{STCD}}\)(ours) & 50 & 1E-03 & 0.170 & 0.047 & 0.313 & - & - \\
			\hline
		\end{tabular}
			\end{adjustbox}
			\begin{adjustbox}{center, margin=0, width=\maxtablewidth}
		\begin{tabular}{|Sc|Sc|Sc|Sc|Sc|Sc|Sc|Sc|}
			\hline
			\multicolumn{8}{|Sc|}{PeMSD8} \\
			\hline
			Models & Batch Size & Learning Rate & $\alpha$1 & $\alpha$2 & $\alpha$3 & $\alpha$ & $\beta$ \\
			
			\hline
			\(L_{\text{RD(L2)}}\) & 50 & 1E-03 & - & - & - & - & 0.905 \\
			\hline
			\(L_{\text{RD(KL)}}\) & 50 & 1E-03 & - & - & - & - & 0.728 \\
			\hline
			\(L_{\text{SKD}}\) & 60 & 1E-02 & - & - & - & 0.005 & - \\
			\hline
			\(L_{\text{ORD}}\)(ours) & 50 & 1E-03 & 0.541 & - & - & - & - \\
			\hline
			\(L_{\text{STCD}}\)(ours) & 50 & 1E-03 & 0.846 & 0.465 & 0.504 & - & - \\
			
			\hline
		\end{tabular}
	\end{adjustbox}
\end{table*}

\begin{table*}[!h]%

	\caption[Hyperparameters of the proposed pruning algorithm]{Hyperparameters of the proposed pruning algorithm}
	\label{tab:4-2}
	
	\hspace*{1cm}
	\begin{adjustbox}{left, margin=0, width=\maxtablewidth}
		\begin{tabular}{|Sc|Sc|Sc|Sc|Sc|Sc|}
			\hline
			\multicolumn{6}{|Sc|}{PeMSD7} \\
			\hline
			Pruned $\%$ & Batch Size & Learning Rate & $\alpha$1 & $\alpha$2 & $\alpha$3 \\
			
			\hline
			97$\%$ & 25 & 1E-03 & 0.746 & 0.445 & 0.020  \\
			\hline
			75$\%$ & 50 & 1E-03 & 0.963 & 0.716 & 0.081  \\
			\hline
			50$\%$ & 50 & 1E-03 & 0.935 & 0.981 & 0.129  \\
			\hline
			25$\%$ & 50 & 1E-03 & 0.971 & 0.234 & 0.684  \\
			
			\hline
		\end{tabular}
		
		\begin{tabular}{|Sc|Sc|Sc|Sc|Sc|Sc|}
			\hline
			\multicolumn{6}{|Sc|}{PeMSD8} \\
			\hline
			Pruned $\%$ & Batch Size & Learning Rate & $\alpha$1 & $\alpha$2 & $\alpha$3 \\
			
			\hline
			97$\%$ & 25 & 1E-03 & 0.099 & 0.091 & 0.531  \\
			
			\hline
			75$\%$ & 50 & 1E-03 & 0.996 & 0.720 & 0.405  \\
			\hline
			50$\%$ & 50 & 1E-03 & 0.946 & 0.516 & 0.094  \\
			\hline
			25$\%$ & 50 & 1E-03 & 0.748 & 0.324 & 0.868  \\
			
			\hline
		\end{tabular}
	\end{adjustbox}
	
\end{table*}

\subsection{Knowledge Distillation Analysis}
\begin{figure*}[htbp]
	
\centering
	\begin{subfigure}{0.8\textwidth}

		\includegraphics[width=1\textwidth]{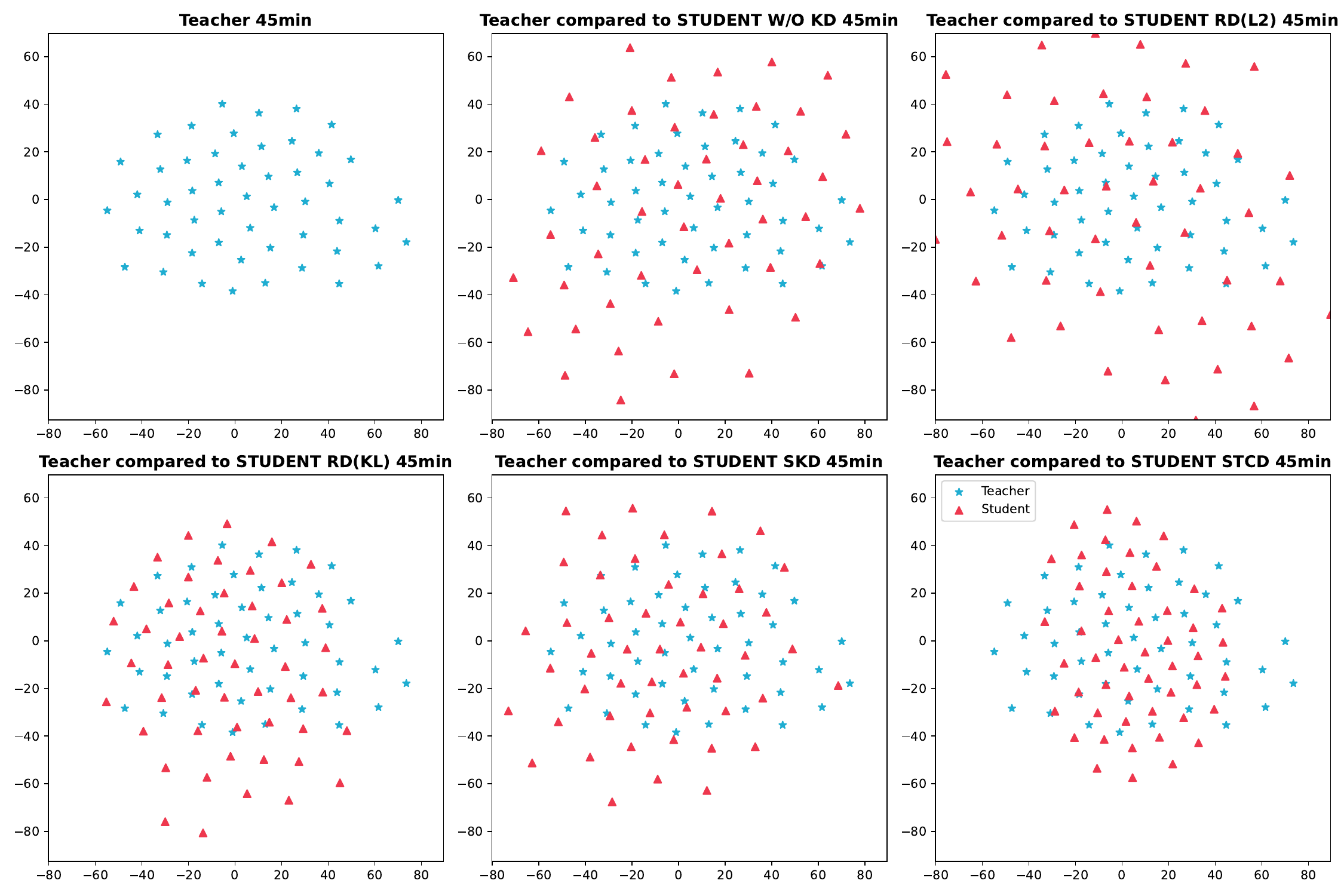}
		\caption{PeMSD7}
		\label{subfig:PeMSD7}
	\end{subfigure}

	\begin{subfigure}{0.8\textwidth}
		\centering
		\includegraphics[width=1\textwidth]{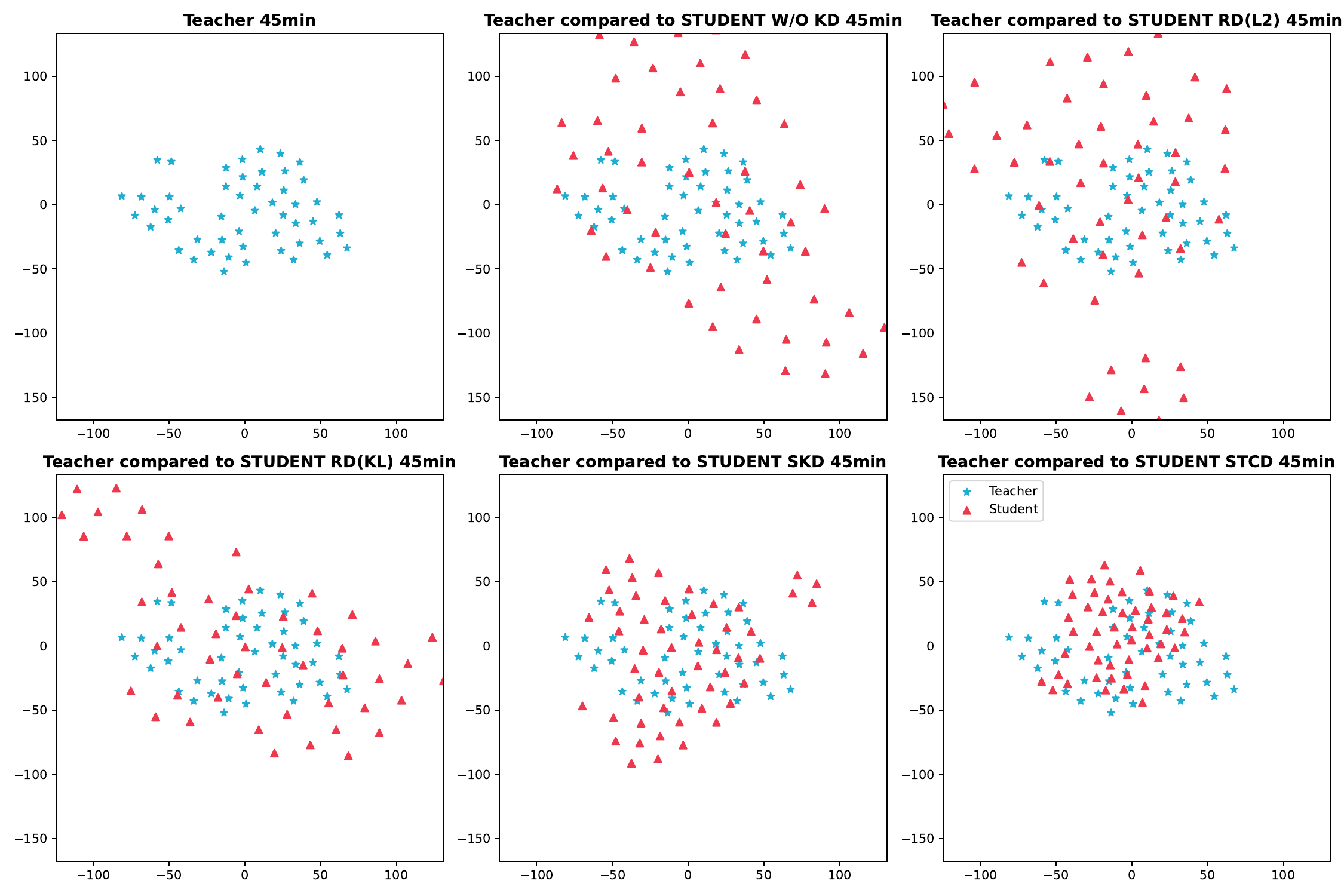}
		\caption{PeMSD8}
		\label{subfig:PeMSD8}
	\end{subfigure}
	
	\caption{Comparison of our spatial-temporal correlation distillation loss function \(L_{\text{STCD}}\) with the \(L_{\text{RD(L2)}}\), \(L_{\text{RD(KL)}}\) and \(L_{\text{SKD}}\) loss functions. Subfigure~\ref{subfig:PeMSD7} shows results for PeMSD7 dataset, and Subfigure~\ref{subfig:PeMSD8} shows for PeMSD8. The left chart in the top row corresponds to the teacher, while the middle chart in the top row pertains to a student without knowledge distillation. In the second row, the right chart represents our loss function, and the other charts indicate different loss functions.}
	\label{fig:3}
\end{figure*}

All knowledge distillation experiments, except for pruning, used the teacher and student models outlined in Table \ref{tab:3-1} and Figure \ref{fig:loss}. These models offer comprehensive information on both networks for easy comparison and analysis. The 'Hidden Blocks Channel' column in the table represents channels for spatial and temporal layers in the two hidden blocks of the ST-GCN network. These channels are visualized in the accompanying Figure. Additionally, 'Test Time (s)' denotes average test time over 100 trials, and 'FLOPS' indicates the network's floating-point operations per second (FLOPS) on a single forward pass. The student to teacher parameter ratio is approximately $3\%$ in both datasets. The student network's execution time has decreased by around 83-84$\%$ compared to the teacher. Both networks have the same number of blocks and layers, with differences in parameter count due to variations in channel numbers per layer. The FLOPS ratio of the student network to the teacher is approximately $4\%$ for the PeMSD7 dataset and $2.5\%$ for the PeMSD8 dataset.
\mbox{}\\We conducted a comparative analysis of our final loss function, \(L_{\text{STCD}}\), against several benchmarks: the baseline student model without knowledge distillation (KD), ARIMA, Historical Average, ASTGCN\cite{ref5, ref60}, and other established approaches \cite{ref11, ref12, ref14, ref13, ref9}. The results are summarized in Table \ref{tab:3-2}. 
\mbox{}\\
In reviewing the table, it is evident that for the PeMSD7 dataset, \(L_{\text{STCD}}\) consistently achieves lower Mean Absolute Error (MAE) and Root Mean Square Error (RMSE) for both 15-minute and 30-minute predictions. Similarly, on the PeMSD8 dataset, \(L_{\text{STCD}}\) demonstrates superior performance across various metrics when compared to \(L_{\text{RD(L2)}}\), \(L_{\text{RD(KL)}}\), and \(L_{\text{SKD}}\), as well as to lightweight models like ASTGCN.	 Figure \ref{fig:3} depict hidden layer representations for 50 test data points, with each point representing an individual instance. Red dots represent the student, while blue dots denote either the teacher or the baseline pruning network (utilized for specific pruning results). Proximity or overlap between red and blue dots serves as an indicator of the model's success. Ideally, effective methods show red and blue dots close to or overlapping each other. Figure \ref{fig:3} illustrates better convergence and pattern imitation of the \(L_{\text{STCD}}\) approach compared to other approaches in both knowledge bases.	

\begin{table*}[!h]%
	\centering
	
	\caption{Comparison of our approach \(L_{\text{STCD}}\) with Historical Average, ARIMA, \(L_{\text{RD(L2)}}\), \(L_{\text{RD(KL)}}\) and \(L_{\text{SKD}}\) on PeMSD7 and PeMSD8}
	\label{tab:3-2}
	\begin{adjustbox}{center, margin=0, width=\maxtablewidthh}
			\begin{tabular}{|Sc|ScScSc|ScScSc|ScScSc|}
		\hline
		\multirow{3}{*}{Models} &
		\multicolumn{9}{|Sc|}{PeMSD7} \\  
		\cmidrule{2-10}
		&\multicolumn{3}{|Sc|}{MAPE} & \multicolumn{3}{Sc|}{MAE} & \multicolumn{3}{Sc|}{RMSE}\\
		\cmidrule{2-10}
		& 15 min & 30 min & 45 min & 15 min & 30 min & 45 min & 15 min & 30 min & 45 min  \\
		\hline
		Teacher & 5.223 & 7.316 & 8.739 & 2.230 & 3.010 & 3.565 & 4.097 & 5.752 & 6.834 \\
		\hline
		
		\text{HA} &  & 10.610 & &  & 4.010 & & & 7.200  &   \\
		\hline
		ARIMA  & 13.400 & 14.010 & 15.010 & 5.570 & 5.940 & 6.270 & 9.000 & 9.220 & 9.430  \\
		\hline	
		ASTGCN  & 7.250 & 8.670 & 9.730 & 2.850 & 3.350 & 3.700 & 5.150 & 6.120 & 6.770 \\
		\hline	
		Student without KD  & 6.423 & 9.685 & 12.298 & 2.666 & 3.868 & 4.799 & 4.649 & 6.938 & 8.610 \\
		\hline
		Student \(L_{\text{RD(L2)}}\) &  6.379 & 9.661 & 12.474 & 2.768 & 4.214 & 5.527 & 4.709 & 7.185 & 9.178 \\
		\hline
		Student \(L_{\text{RD(KL)}}\) &  6.411 &  9.527 & 11.894 & 2.700 & 3.938 & 4.918 & 4.672 & 6.984 & 8.657  \\
		\hline
		Student \(L_{\text{SKD}}\) & 6.282 & 9.503 & 11.902 & 2.762 & 4.092 & 5.101 & 4.735 & 7.161 & 8.923  \\
		\hline			
		Student \(L_{\text{STCD}}\)(ours) & 6.078 & 9.043 & 11.488 & 2.615 & 3.776 & 4.754 & 4.537 & 6.678 & 8.344 \\
		\hline
	\end{tabular}
	\end{adjustbox}
		\begin{adjustbox}{center, margin=0, width=\maxtablewidthh}

			\begin{tabular}{|Sc|ScScSc|ScScSc|ScScSc|}
			\hline
			\multirow{3}{*}{Models} &
			\multicolumn{9}{|Sc|}{PeMSD8} \\  
			\cmidrule{2-10}
			&\multicolumn{3}{|Sc|}{MAPE} & \multicolumn{3}{Sc|}{MAE} & \multicolumn{3}{Sc|}{RMSE}\\
			\cmidrule{2-10}
			& 15 min & 30 min & 45 min & 15 min & 30 min & 45 min & 15 min & 30 min & 45 min  \\
			\hline
			Teacher & 2.293 & 3.239 & 3.925 & 1.211 & 1.665 & 2.031 & 2.524 & 3.501 & 4.081 \\
			\hline
			
			\text{HA} &  & 3.940 & &  & 1.980 & &  & 4.110 & \\
			\hline
			ARIMA & 5.110 & 5.210 & 5.460 & 1.900 & 2.120 & 2.430 & 4.870 & 5.240 & 5.630 \\
			\hline	
			ASTGCN & 3.160 & 3.590 & 3.980 & 1.490 & 1.670 & 1.81 & 3.180 & 3.690 & 3.920 \\
			\hline	
			Student without KD & 2.967 & 4.035 & 4.734 & 1.472 & 1.956 & 2.294 & 2.988 & 4.090 & 4.706 \\
			\hline
			Student \(L_{\text{RD(L2)}}\) & 2.812 & 3.976 & 4.908 & 1.426 & 1.953 & 2.387 & 2.839 & 3.978 & 4.733 \\
			\hline
			Student \(L_{\text{RD(KL)}}\) & 2.964 &  4.179 & 5.057 & 1.507 & 2.125 & 2.575 & 2.895 & 3.971 & 4.664 \\
			\hline
			Student \(L_{\text{SKD}}\) & 2.726 & 3.819 & 4.553 & 1.409 & 1.944 & 2.317 & 2.909 & 4.139 & 4.894 \\
			\hline			
			
			Student \(L_{\text{STCD}}\)(ours) &  2.491 & 3.375 & 4.067 & 1.281 & 1.719 & 2.052 & 2.716 & 3.707 & 4.385 \\
			\hline
		\end{tabular}
		
	\end{adjustbox}
\end{table*} 

\mbox{}\\
We also conducted an ablation study on our final loss function, $L_{\text{STCD}}$; response-based distillation, $L_{\text{ORD}}$; and other variants, including $L_{\text{TCD}}$, $L_{\text{SCD}}$, along with the baseline Student without KD (student trained without KD), and the findings are outlined in Table \ref{tab:3-3}.
Additionally, we report the ratio of attention to teacher output instead of the dataset target in Table \ref{tab:3-0}.
\begin{table*}[!h]%
	\centering
	
	\caption{Ablation study for our \(L_{\text{STCD}}\) loss function on PeMSD7 and PeMSD8}
	\label{tab:3-3}
	\begin{adjustbox}{center, margin=0, width=\maxtablewidthh}
		\begin{tabular}{|Sc|ScScSc|ScScSc|ScScSc|}
			\hline
			\multirow{3}{*}{Models} &
			\multicolumn{9}{|Sc|}{PeMSD7} \\  
			\cmidrule{2-10}
			&\multicolumn{3}{|Sc|}{MAPE} & \multicolumn{3}{Sc|}{MAE} & \multicolumn{3}{Sc|}{RMSE}\\
			\cmidrule{2-10}
			& 15 min & 30 min & 45 min & 15 min & 30 min & 45 min & 15 min & 30 min & 45 min \\
			\hline
			Teacher & 5.223 & 7.316 & 8.739 & 2.230 & 3.010 & 3.565 & 4.097 & 5.752 & 6.834  \\
			\hline
			Student without KD  & 6.423 & 9.685 & 12.298 & 2.666 & 3.868 & 4.799 & 4.649 & 6.938 & 8.610  \\
			\hline
			Student \(L_{\text{ORD}}\) & 6.411 & 9.516 & 12.104 & 2.762 & 4.091 & 5.221 & 4.645 & 6.847 & 8.569 \\
			\hline
			Student \(L_{\text{TCD}}\)  & 6.320 & 9.411 & 11.667 & 2.730 & 3.966 & 4.853 & 4.678 & 6.899 & 8.512 \\
			\hline
			Student \(L_{\text{SCD}}\)  & 6.326 & 9.193 & 11.380 & 2.743 & 3.957 & 4.866 &4.645& 6.853 & 8.476  \\
			\hline
			Student \(L_{\text{STCD}}\) & 6.078 & 9.043 & 11.488 & 2.615 & 3.776 & 4.754 & 4.537 & 6.678 & 8.344 \\
			\hline
		\end{tabular}
		
	\end{adjustbox}
		\begin{adjustbox}{center, margin=0, width=\maxtablewidthh}
		\begin{tabular}{|Sc|ScScSc|ScScSc|ScScSc|}
			\hline
			\multirow{3}{*}{Models} &
			\multicolumn{9}{|Sc|}{PeMSD8} \\  
			\cmidrule{2-10}
			&\multicolumn{3}{|Sc|}{MAPE} & \multicolumn{3}{Sc|}{MAE} & \multicolumn{3}{Sc|}{RMSE}\\
			\cmidrule{2-10}
			& 15 min & 30 min & 45 min & 15 min & 30 min & 45 min & 15 min & 30 min & 45 min \\
			\hline
			Teacher  & 2.293 & 3.239 & 3.925 & 1.211 & 1.665 & 2.031 & 2.524 & 3.501 & 4.081 \\
			\hline
			Student without KD & 2.967 & 4.035 & 4.734 & 1.472 & 1.956 & 2.294 & 2.988 & 4.090 & 4.706 \\
			\hline
			Student \(L_{\text{ORD}}\)  & 2.661 & 3.717 & 4.553 & 1.363 & 1.885 & 2.285 & 2.788 & 3.862 & 4.573 \\
			\hline
			Student \(L_{\text{TCD}}\)  & 2.509 & 3.497 & 4.215 & 1.296 & 1.747 & 2.063 & 2.665 & 3.716 & 4.406 \\
			\hline
			Student \(L_{\text{SCD}}\)  & 2.619 & 3.680 & 4.457 & 1.355 & 1.850 & 2.214 & 2.722 & 3.778 & 4.478 \\
			\hline
			Student \(L_{\text{STCD}}\) & 2.491 & 3.375 & 4.067 & 1.281 & 1.719 & 2.052 & 2.716 & 3.707 & 4.385 \\
			\hline
		\end{tabular}
		
	\end{adjustbox}
\end{table*} 

\begin{table*}[!h]%
	\centering
	
	\caption{Ratios of attention to teacher prediction to all training data}
	\label{tab:3-0}
	\begin{center}
		\begin{adjustbox}{center, margin=0, width=\maxtablewidth}
			\begin{tabular}{|Sc|>{\centering\arraybackslash}p{3cm}|>{\centering\arraybackslash}p{3cm}|}
				\hline
				\multirow{2}{*}{Models} & \multicolumn{2}{Sc|}{Ratios of Attention to Teacher Prediction}\\ \cmidrule{2-3}
				& PeMSD7 &  PeMSD8 \\
				\hline
				Student \(L_{\text{ORD}}\) &  3.009\% &  0.749\%  \\
				\hline
				Student \(L_{\text{SCD}}\)  & 1.377\% & 0.046\%  \\
				\hline
				Student \(L_{\text{TCD}}\) &  64.637\% & 0.503\%  \\
				\hline
				Student \(L_{\text{STCD}}\) &  15.844\% &  0.103\%  \\
				\hline
			\end{tabular}
		\end{adjustbox}
	\end{center}
\end{table*}

\subsection{Pruning and Fine-Tuning Analysis}
\begin{figure*}[t]
	
	\begin{subfigure}{0.3\textwidth}
		
		\includegraphics[width=1.8\textwidth]{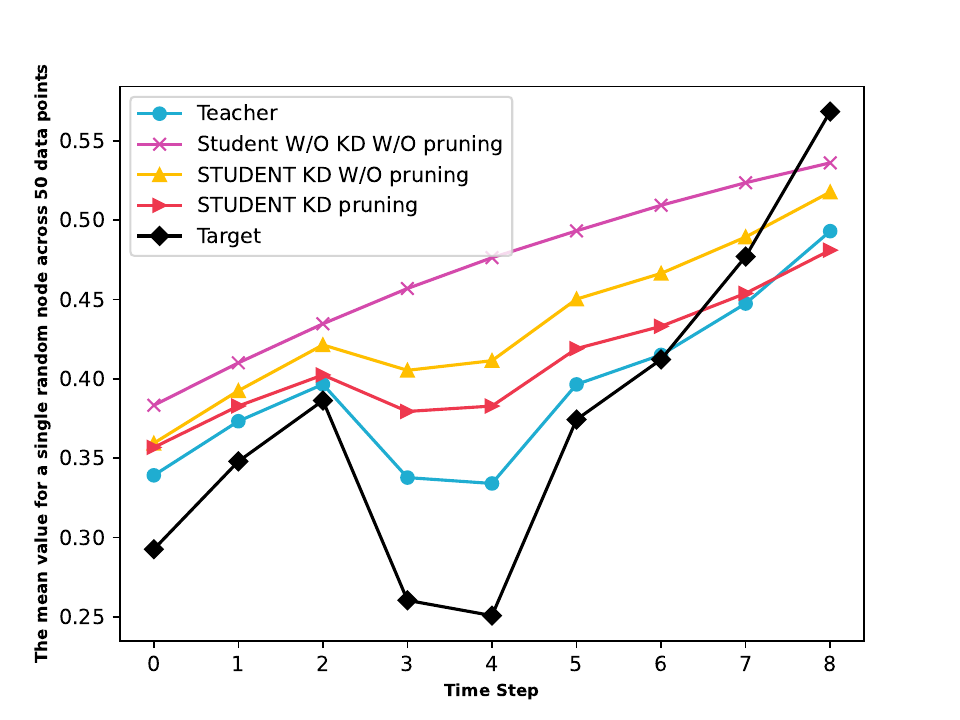}
		\caption{PeMSD7}
		\label{subfig:PeMSD73}
	\end{subfigure}
	\qquad
	\hspace*{2cm}
	\begin{subfigure}{0.3\textwidth}
		
		\includegraphics[width=1.8\textwidth]{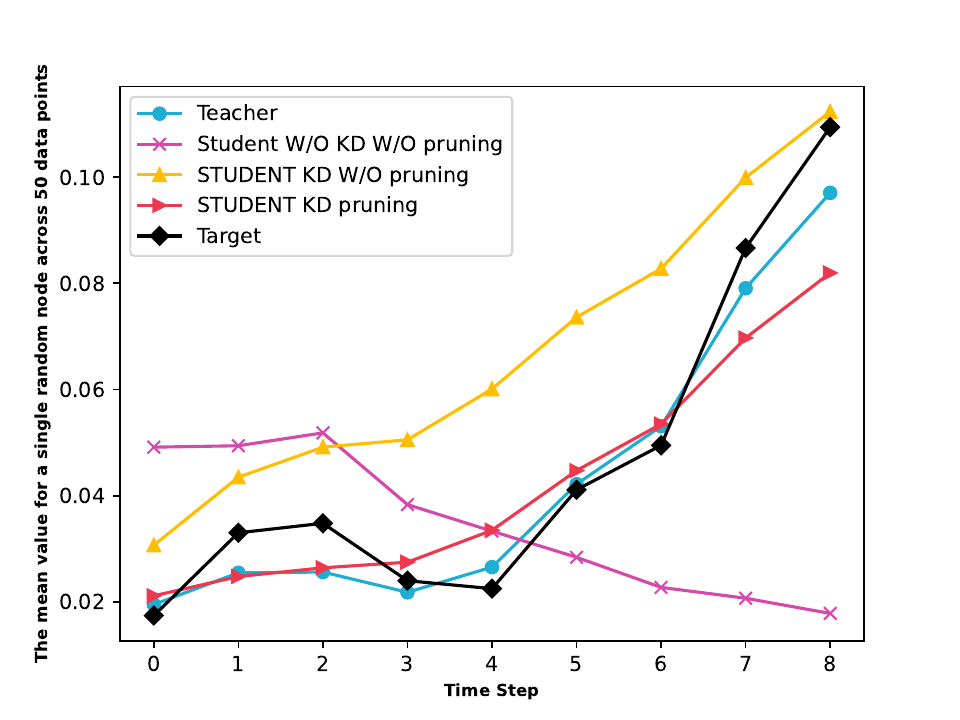}
		\caption{PeMSD8}
		\label{subfig:PeMSD83}
	\end{subfigure}
	\caption[The results highlight the impact of employing the pruning algorithm, demonstrating the average predicted value for a randomly selected node based on 50 data points.]{The results highlight the impact of employing the pruning algorithm, demonstrating the average predicted value for a randomly selected node based on 50 data points.}
	\label{fig:7}
\end{figure*}

In this section, the goal is to showcase that a consciously pruned student network, guided by Algorithm 1, exhibits superior learning capabilities compared to a predetermined student network used in knowledge distillation. The experiment's outcomes are summarized in Table \ref{tab:3-4} and Figure \ref{fig:7}. Table \ref{tab:3-4} presents results comparing the performance of the student network with and without pruning through knowledge distillation. Looking at the PeMSD7 dataset in the table, the teacher model, using all of its parameters, achieved an RMSE of 4.097 for a 15-minute prediction. On the other hand, the student model without the pruning algorithm, which retained about 3$\%$ of the parameters, had a higher RMSE of 4.537 for the same prediction period. However, when we applied the pruning algorithm to the student model (also retaining 3$\%$ of parameters), the RMSE improved to 4.398. These results vividly demonstrate that the pruning algorithm significantly contributes to enhancing knowledge distillation, showcasing instances where pruning leads to better model performance compared to the non-pruned counterparts.

\begin{table*}[!h]%
	\centering
	\caption{The results showcase the impact of applying the pruning algorithm in enhancing knowledge distillation.}
	\label{tab:3-4}
	\begin{adjustbox}{center, margin=0, width=\maxtablewidth}
		\begin{tabular}{|Sc|Sc|ScScSc|ScScSc|ScScSc|}
			\hline
			\multicolumn{11}{|Sc|}{PeMSD7} \\
			\hline
			\multirow{2}{*}{Models} & \multirow{2}{*}{Kept parameter $\%$} & \multicolumn{3}{Sc|}{MAPE} & \multicolumn{3}{Sc|}{MAE} & \multicolumn{3}{Sc|}{RMSE} \\
			\cmidrule{3-11}
			& &15 min & 30 min & 45 min & 15 min & 30 min & 45 min & 15 min & 30 min & 45 min \\
			\hline
			
			Teacher  & $100\%$& 5.223 & 7.316 & 8.739 & 2.230 & 3.010 & 3.565 & 4.097 & 5.752 & 6.834 \\
			\hline
			Student without pruning algorithm & $~3\%$& 6.078 & 9.043 & 11.488 & 2.615 & 3.776 & 4.754 & 4.537 & 6.678 & 8.344 \\
			\hline
			Student with pruning algorithm  & $~3\%$& 5.691 & 8.410 & 10.520 & 2.470 & 3.366 & 4.131 & 4.264 & 6.152 & 7.564\\
			\hline
			
		\end{tabular}
	\end{adjustbox}
	\begin{adjustbox}{center, margin=0, width=\maxtablewidth}
		\begin{tabular}{|Sc|Sc|ScScSc|ScScSc|ScScSc|}
			\hline
			\multicolumn{11}{|Sc|}{PeMSD8} \\
			\hline
			\multirow{2}{*}{Models} & \multirow{2}{*}{Kept parameter $\%$} & \multicolumn{3}{Sc|}{MAPE} & \multicolumn{3}{Sc|}{MAE} & \multicolumn{3}{Sc|}{RMSE} \\
			\cmidrule{3-11}
			& &15 min & 30 min & 45 min & 15 min & 30 min & 45 min & 15 min & 30 min & 45 min \\
			\hline
			Teacher  & $100\%$& 2.293 & 3.239 & 3.925 & 1.211 & 1.665 & 2.031 & 2.524 & 3.501 & 4.081 \\
			\hline
			Student without pruning algorithm & $~3\%$& 2.491 & 3.375 & 4.067 & 1.281 & 1.719 & 2.052 & 2.716 & 3.707 & 4.385 \\
			\hline
			Student with pruning algorithm  & $~3\%$& 2.379 & 3.307 & 3.961 & 1.248 & 1.679 & 1.990 & 2.597 & 3.594 & 4.193 \\
			\hline
		\end{tabular}
	\end{adjustbox}
\end{table*}

\mbox{}\\
Previous research has confirmed that incorporating knowledge distillation enhances accuracy while utilizing a more straightforward model. Building upon this, our current study extends the concept, illustrating that Algorithm 1 yields a student architecture that performs better during training with KD. We compare Algorithm 1 with \cite{ref8} using a base pruning model from Table \ref{tab:3-1} (second row). The results depicted in Table \ref{tab:3-5} underscore that our pruning algorithm consistently outperforms the alternative across various pruning percentages. Overall, the findings affirm that our introduced pruning algorithm plays a pivotal role in enhancing learning through knowledge distillation, significantly improving neural network accuracy by employing fewer parameters and simpler structures.
\begin{table*}[t]%
	\centering
	\caption{Comparison of the performance between Algorithm 1 and the pruning algorithm that does not utilize knowledge distillation.}
	\label{tab:3-5}
	\begin{adjustbox}{center, margin=0, width=\maxtablewidth}
		\begin{tabular}{|Sc|Sc|ScScSc|ScScSc|ScScSc|}
			\hline
			\multicolumn{11}{|Sc|}{PeMSD7} \\
			\hline
			\multirow{2}{*}{Models} & \multirow{2}{*}{Kept parameters  $\%$} & \multicolumn{3}{Sc|}{MAPE} & \multicolumn{3}{Sc|}{MAE} & \multicolumn{3}{Sc|}{RMSE} \\
			\cmidrule{3-11}
			& &15 min & 30 min & 45 min & 15 min & 30 min & 45 min & 15 min & 30 min & 45 min \\
			\hline
			
			Base Model  & $100\%$& 5.512 & 8.004 & 9.988 & 2.321 & 3.216 & 3.896 & 4.177 & 6.006 & 7.341 \\
			\hline
			Traditional pruning & $75\%$& 6.185 & 8.981 & 11.060 & 2.721 & 3.889 & 4.728 & 4.682 & 6.947 & 8.579 \\
			\hline
			Pruned with \(L_{\text{STCD}}\)(ours) & $75\%$& 5.950 & 8.654 & 10.906 & 2.544 & 3.705 & 4.751 & 4.406 & 6.424 & 8.014 \\
			\hline
			Traditional pruning & $50\%$& 6.571 & 9.735 & 12.114 & 2.975 & 4.389 & 5.418 & 4.858 & 7.255 & 8.971 \\
			\hline
			Pruned with \(L_{\text{STCD}}\)(ours) & $50\%$& 6.232 & 9.185 & 11.419 & 2.584 & 3.773 & 4.723 & 4.524 & 6.773 & 8.448 \\
			\hline
			Traditional pruning & $25\%$& 7.090 & 11.746 & 15.940 & 2.925 & 4.635 & 6.034 & 4.943 & 7.759 & 9.967 \\
			\hline
			Pruned with \(L_{\text{STCD}}\)(ours) & $25\%$& 6.275 & 9.436 & 12.261 & 2.644 & 3.783 & 4.708 & 4.586 & 6.680 & 8.270 \\
			\hline
		\end{tabular}
	\end{adjustbox}
	\begin{adjustbox}{center, margin=0, width=\maxtablewidth}
		\begin{tabular}{|Sc|Sc|ScScSc|ScScSc|ScScSc|}
			\hline
			\multicolumn{11}{|Sc|}{PeMSD8} \\
			\hline
			\multirow{2}{*}{Models} & \multirow{2}{*}{Kept parameters  $\%$} & \multicolumn{3}{Sc|}{MAPE} & \multicolumn{3}{Sc|}{MAE} & \multicolumn{3}{Sc|}{RMSE} \\
			\cmidrule{3-11}
			& &15 min & 30 min & 45 min & 15 min & 30 min & 45 min & 15 min & 30 min & 45 min \\
			\hline
			Base Model  & $100\%$& 2.206 & 3.028 & 3.677 & 1.187 & 1.592 & 1.924 & 2.479 & 3.434 & 4.084 \\
			\hline
			Traditional pruning & $75\%$& 2.438 & 3.459 & 4.338 & 1.279 & 1.781 & 2.227 & 2.615 & 3.644 & 4.359 \\
			\hline
			Pruned with \(L_{\text{STCD}}\)(ours) & $75\%$ & 2.427 & 3.200 & 3.782 & 1.236 & 1.630 & 1.927 & 2.672 & 3.645 & 4.303 \\
			\hline
			Traditional pruning & $50\%$& 2.456 & 3.514 & 4.449 & 1.322 & 1.890 & 2.397 & 2.648 & 3.797 & 4.685 \\
			\hline
			Pruned with \(L_{\text{STCD}}\)(ours) & $50\%$& 2.424 & 3.310 & 3.988 & 1.261 & 1.693 & 2.029 & 2.605 & 3.550 & 4.169 \\
			\hline
			Traditional pruning & $25\%$& 2.530 & 3.503 & 4.275 & 1.335 & 1.823 & 2.225 & 2.715 & 3.810 & 4.596 \\
			\hline
			Pruned with \(L_{\text{STCD}}\)(ours) & $25\%$& 2.348 & 3.192 & 3.788 & 1.228 & 1.627 & 1.911 & 2.595 & 3.577 & 4.216 \\
			\hline
		\end{tabular}
	\end{adjustbox}
\end{table*}

\section{Conclusion and Future Works}
\subsection{Conclusion}
We addressed the critical challenge of predicting traffic conditions to reduce transportation time. Our chosen methodology involved leveraging the spatio-temporal graph convolutional network (ST-GCN) for processing traffic prediction data, and we proposed a solution to improve the execution time of ST-GCNs. This involved introducing a novel approach that employed knowledge distillation to train a smaller network (referred to as the student) using distilled data from a more complex network (referred to as the teacher), all while maintaining a high level of prediction accuracy. Additionally, we tackled the challenge of determining the architecture of the student network within the knowledge distillation process, employing Algorithm 1. Our findings demonstrated that a student network derived through Algorithm 1 exhibited enhanced performance during the knowledge distillation training process. Furthermore, by utilizing a simpler model in knowledge distillation, as opposed to the teacher model, as the base model in pruning, we illustrated that the performance of our algorithm was not necessarily contingent on the high accuracy of the base model.
We evaluated our proposed concepts using two real-world datasets, PeMSD7 and PeMSD8, demonstrating the effectiveness of our approach in predicting traffic conditions. Overall, our research contributed to the field by presenting a methodology to optimize the execution time of spatio-temporal graph convolutional networks for traffic prediction without compromising accuracy.

\subsection{Future Works}
As we progress, we have the opportunity to explore our cost function  \(L_{\text{STCD}}\) and Algorithm 1 on alternative spatiotemporal graph neural networks, such as ASTGCN. Furthermore, considering the method proposed in \cite{ref15} and \cite{ref10}, we can explore whether using multiple teachers in Algorithm 1 and knowledge distillation with \(L_{\text{STCD}}\) can lead to a model with higher accuracy. Alternatively, we can develop a model capable of executing the tasks performed by all its teachers with increased accuracy and reduced execution time compared to the model in \cite{ref10}. It is possible to enhance the concept of knowledge distillation based on the cost function \(L_{\text{STCD}}\) by incorporating relationship-based approaches. This improvement allows for the consideration of student and teacher networks with varying numbers of layers.\\
Additionally, improving the interpretability of distilled models can significantly increase trust and provide deeper insights into traffic patterns. Understanding how the student network makes its predictions can help identify potential improvements and uncover new traffic dynamics. For example, employing techniques like the ones proposed in \cite{ref49}, which focus on enhancing the transparency of deep learning models, could be beneficial for our traffic prediction models.\\	
Finally, developing advanced pruning techniques that consider structural properties of the network can lead to more efficient and compact models without sacrificing accuracy. By focusing on the importance of connections within the network, as seen in recent methods such as those described in \cite{ref50}, we can achieve significant reductions in model complexity while maintaining high performance. This approach aligns with our goal of optimizing the student network for real-time traffic prediction on resource-constrained devices.\\
 In summary, future work can expand on our current framework by integrating adaptive learning mechanisms, improving model interpretability, and utilizing advanced pruning techniques, all of which will contribute to creating more robust and efficient traffic prediction models.

\section*{Declaration of Competing Interest}
The authors declare that they have no known competing financial interests or personal relationships that could have appeared to influence the work reported in this paper.

\section*{Acknowledgements}
This research did not receive any specific grant from funding agencies in the public, commercial, or not-for-profit sectors.

\section*{Data availability}
Data will be made available on request. 

\section*{Declaration of Generative AI and AI-assisted technologies in the writing process}
During the preparation of this work the author(s) used ChatGPT 3.5 in order to refine language. After using this tool/service, the author(s) reviewed and edited the content as needed and take(s) full responsibility for the content of the publication.

\end{document}